\documentclass[runningheads]{llncs}
\usepackage{graphicx}
\usepackage{setspace}
\usepackage{booktabs}
\usepackage{array}
\usepackage{threeparttable}
\usepackage{gensymb}
\usepackage{caption}
\usepackage{subfigure}
\usepackage{setspace}
\usepackage{cite}
\usepackage{makecell}
\usepackage{multirow}
\usepackage{tabularx}
\usepackage{color,soul}
\usepackage{amssymb}
\usepackage{hyperref}

\begin{document}
\title{Semi-supervised Contrastive Learning for Label-efficient Medical Image Segmentation}
\author{Xinrong Hu\inst{1}
\and Dewen Zeng\inst{1}
\and Xiaowei Xu\inst{2}
\and Yiyu Shi\inst{1}}

\authorrunning{X. Hu et al.}

\institute{University of Notre Dame, Notre Dame IN 46556, USA\\ 
\email{\{xhu7, dzeng2, yshi4\}@nd.edu}
\and
Guangdong Provincial People's Hospital, Guangdong 510000, China}
\titlerunning{Semi-supervised Contrastive Learning}
%
%
\maketitle              
\begin{abstract}
The success of deep learning methods in medical image segmentation tasks heavily depends on a large amount of labeled data to supervise the training. 
On the other hand, the annotation of biomedical images requires domain knowledge and can be 
laborious. Recently, contrastive learning has demonstrated great potential in learning latent representation of images even without any label. Existing works have explored 
its application to biomedical image segmentation where only a small portion of data is 
labeled, through a pre-training phase based on self-supervised contrastive learning without using any labels followed by a supervised fine-tuning phase on the labeled portion of data only. 
In this paper, we establish that 
by including the limited label information in the pre-training phase, it is 
possible to boost the performance of contrastive learning. 
We propose a supervised local contrastive loss that leverages limited pixel-wise annotation to force pixels with the same label to gather around in the embedding space. 
Such loss needs pixel-wise computation which can be expensive for large images, and 
we further propose two strategies, downsampling and block division, to address the issue. 
We evaluate our methods on two public biomedical image datasets of different modalities. 
With different amounts of labeled data, our methods consistently outperform the state-of-the-art contrast-based methods and other semi-supervised learning techniques. 
We also publish our code at: \href{https://github.com/xhu248/semi_cotrast_seg}{https://github.com/xhu248/semi\_cotrast\_seg}

\end{abstract}

\section{Introduction}
\label{sec:introdcution}
Accurate semantic segmentation result is of great value to medical application, which provides physicians with the anatomical structural information for disease diagnosis and treatment.
With the emergence of convolutional neural network, supervised deep learning has achieved state-of-the-art performance in many biomedical image segmentation tasks, including different organs and different modalities~\cite{ronneberger2015u, chen2017deeplab, milletari2016v, li2018h}.
These methods all need abundant labeled data with the class of every pixel or voxel known to guide the training. 
However, it is difficult to collect densely labeled medical images, because labeling medical images require domain-specific knowledge and pixel-wise annotations can be time-consuming. 

To deal with this issue, various methods have been proposed.
One branch of them is data augmentation, which expands labeled dataset by generative adversarial networks (GAN) \cite{goodfellow2014generative} for data synthesis \cite{chaitanya2019semi, bowles2018gan, costa2017end} or by simple linear combination\cite{zhang2017mixup}. 
However, data augmentation only leads to limited improvement of segmentation performance due to the defect of artificially generated images and labels.
As a better alternative, semi-supervised learning based approaches \cite{bai2017semi, zhang2017deep, li2020transformation, li2020shape} can utilize both labeled data and unlabeled data efficiently. 
Recently, contrastive learning\cite{chen2020simple, he2020momentum}, by forcing the embedding features of similar images to be close in the latent space and those of dissimilar ones to be apart, achieved state-of-the-art in self-supervised classification problems.
The powerful ability of extracting features that are useful for downstream tasks from unlabeled data makes it a great candidate for label efficient learning.

However, most existing contrastive learning methods target image classification tasks. Semantic segmentation, on the other hand, requires pixel-wise classification.
\cite{chaitanya2020contrastive, zeng2021positional} first attempted at using contrastive learning to improve segmentation accuracy when only part of the data is labeled. For \cite{chaitanya2020contrastive}, a two-step self-supervised contrastive learning scheme is used to learn both global and local features from unlabeled data in the pre-training phase,. 
Specifically, it first projects a 2D slice to latent space with the encoder path only and computes a global contrastive loss, similar to what has been used for image classification problem.
Then based on the trained encoder, decoder blocks are added and the intermediate feature maps are used to perform local contrastive learning at pixel level.
Only a small number of points are selected from fixed positions in the feature maps.
Next, in the fine-tuning phase the pre-trained model from contrastive learning is trained with supervision on the labeled part of the data. 
In such a framework, the label information is never utilized in the pre-training phase. Yet intuitively, by including such information to provide certain supervision, the features can be better extracted by contrastive learning. 


Following this philosophy, in this work we propose a semi-supervised framework consisting of self-supervised global contrast and supervised local contrast to take advantage of the available labels.
Compared with the unsupervised local contrast in the literature, the supervised one can better enforce the similarity between features within the same class and discriminate those that are not.
In addition, we use two strategies, downsampling and block division, to address the high computation complexity associated with the supervised local contrastive loss. 
We conduct experiments on two public medical image segmentation datasets 
of different modalities.  
We find that, for different amount of labeled data, our framework can always generate segmentation results with higher Dice than the state-of-the-art methods.


\section{Methods}
\label{sec:methods}

\begin{figure}
    \centering
     \includegraphics[width= 0.8\linewidth]{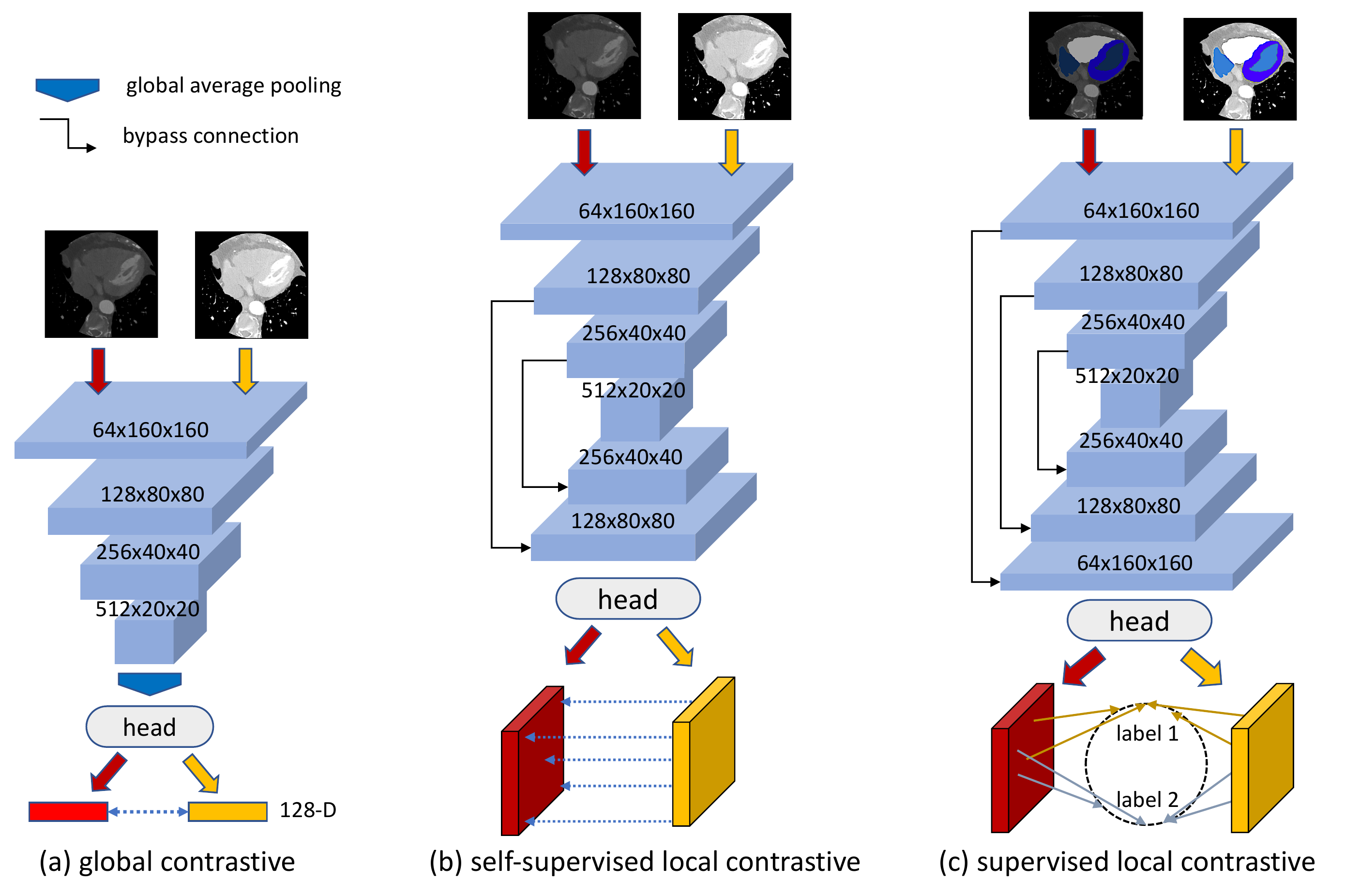}
    \caption{Illustration of (a) self-supervised global contrastive loss, (b) self-supervised local contrastive loss used in existing works, and (c) our proposed supervised local contrastive loss.}
    \label{fig:workflow}
\end{figure}

\subsection{Overview of the Semi-supervised Framework}
The architecture backbone of a deep segmentation network usually consists of two parts, the encoder $E$ and the decoder $D$.
The encoder comprises several downsampling blocks to extract features from input images and the decoder path deploys multiple upsampling blocks to restore the resolution for pixel-wise prediction.
In the pre-training stage, our framework contains a self-supervised global contrastive 
learning step which only uses unlabeled data to train $E$ and captures image-level features, followed by a supervised local contrastive learning step 
which uses the limited labeled data to further train both $E$ and $D$ 
and captures pixel-level features.

The global contrastive learning is similar to what is used in the literature (e.g. SimCLR \cite{chen2020simple}) and for the sake of completeness we briefly describe it here with an 
illustration in Fig.~\ref{fig:workflow}(a). 
Given an input batch $B = \{x_{1}, x_{2}, ..., x_{b}\}$, in which $x_{i}$ represents an input 2D slice,
random combinations of image transformations $aug(\cdot)$ are performed twice on each image in $B$ to form a pair of augmented images, same as the existing contrastive learning methods.
These augmented images form an augmented image set A.
Let $a_{i}, i\in I =\{1...2b\}$ represent an augmented image in $A$ and let j(i) be the index of the other image in $A$ that is derived from the same image in $B$ as $a_{i}$, i.e., $a_{i}$ and 
$a_{j(i)}$ is an augmented pair.
The global contrastive loss then takes the following form
\begin{equation}
\label{equation1}
    L_{g} = -\frac{1}{\left | A \right |}\sum_{i\in I}log\frac{exp(z_{i}\cdot z_{j(i)}/\tau )}{\sum_{k\in I -\{i\} }exp(z_{i}\cdot z_{k})/\tau)},
\end{equation}
where $z_{i}$ is the normalized output feature by adding a header $h_{1}(\cdot)$ to encoder $E$. The header is usually a multi-layer perceptron that enhances the representing ability of extracted features from $E$. Mathematically put, $z_{i} = |h_{1}(E(a_{i}))|$ and $z_{j(i)} = |h_{1}(E(a_{j(i)}))|$. 
$\tau$ is the temperature constant.

Only teaching the encoder to extract image-level features is not enough, since segmentation requires the prediction of all pixels in the input image.
After the global contrastive learning, we attach the decoder path $D$ to the encoder and retrain the whole network with supervised local contrastive loss. This is where our work differs from  existing works and the details will be described in the following section.
The two-step pre-training stage will eventually generate a model that can be used as an initialization for the subsequent fine-tuning stage on the labeled data to boost segmentation accuracy.

\subsection{Supervised Local Contrastive learning}
The goal of local contrastive learning is to train the decoder to extract distinctive local representations.
Let $f^l(\tilde{x_{i}})=h_{2}(D_{l}(E(a_{i})))$ be the output feature map of the $l$-th uppermost decoder block $D_{l}$ for an augmented input $a_{i}$, where $h_{2}$ is a two-layer point-wise convolution. 
For feature map $f(a_{i})$, the local contrastive loss is defined as
\begin{equation}
\label{equation2}
    loss(a_{i}) = -\frac{1}{\left | \Omega \right |}\sum_{(u, v) \in \Omega}\frac{1}{\left | P(u, v) \right |}log\frac{\sum_{(u_{p}, v_{p}) \in P(u, v)}exp(f^l_{u ,v}\cdot f^l_{u_{p}, v_{p}}/\tau )}{\sum_{({u}', {v}') \in N(u, v)}exp(f^l_{u ,v}\cdot f^l_{{u}', {p}'}/\tau )}
\end{equation}
$f^{l}_{u,v}$ $\in$ $\mathbb{R}^{c}$, where $c$ is the channel number, stands for the feature at the $u$th column and $v$th row of the feature map.
$\Omega$ is the set of points in $f_{u,v}$ that are used to compute the loss.
$P(u,v)$ and $N(u, v)$ denote the positive set and negative set of $f_{u,v}(a_{i})$, respectively. 
Then the overall local contrastive loss is
\begin{equation}
    L_{l} = \frac{1}{\left | A \right |}\sum_{a_{i} \in A} loss(a_{i})
\end{equation}

In self-supervised setting, \cite{chaitanya2020contrastive} only selects 9 or 13 points from the feature map as illustrated by Fig.~\ref{fig:sample}(b). 
Without annotation information, $P(u,v)$ only contains points from the same position at the feature maps of the paired images, that is  $f_{u,v}(a_{j(i)})$, and the negative set is the union of $\Omega$ of all images in $A$ minus the positive set. 

According to SimCLR, when applying contrastive learning to image classification tasks, the data augmentation that boosts the accuracy most is the spatial transformation, such as 
crop and cutout. 
Yet in self-supervised setting, the transformation $aug(\cdot)$ can only include intensity transformation such as color distort, Gaussian noise and Gaussian blur. 
It is because after spatial transformation, features at the same position of $f(a_i$ and $f(a_{j(i)})$ could originate from different points or areas but would still be treated as similar when calculating (\ref{equation2}). Moreover, without any supervision, chances are that the selected points in $\Omega$ might all correspond to the background area, in which case the model should fail to effectively learn local features. 

Inspired by \cite{khosla2020supervised}, since the dataset contains limited labeled data anyway to be used in the  
fine-tuning stage, we can make use of it in the pre-training stage through a supervised local contrastive learning.
Observing that the last point-wise convolution in U-Net functions similarly as the fully connected layer of classification network, we view the feature maps $f^{1} \in \mathbb{R}^{c\times h \times h}$ just before the 1x1 convolutional layer as the projections of all pixels into the latent space, where $c$ equals the channel number that is also the the dimension of latent space and $h$ is the input image dimension.
Given a point (u, v) in the feature map, the definition of positive set $P(u, v)$ is all features in the feature maps of $A$ that share the same annotation as (u, v).
The negative set is then all features in the feature maps of $A$ that are labeled as different classes.
In addition, two embedding features of background pixels make up the majority of positive pairs and the subsequent segmentation task benefits less from comparing features as these pairs. Hence, for supervised setting, we change the definition of $\Omega$ to only contain points with non-background annotation. 
Please note that background piexls are still included in the negative set $N(u, v)$.
Then the supervised contrastive loss takes the same form as (\ref{equation2}), except for the new meaning of $\Omega, P(u,v), N(u,v)$ as discussed above. 
With the guidance of labels, supervised contrastive loss provides additional information on the similarity of features derived from the same class and dissimilarity of inter-class features.

\subsection{Downsampling and Block Division}
\begin{figure}
    \centering
    \includegraphics[width=1\linewidth]{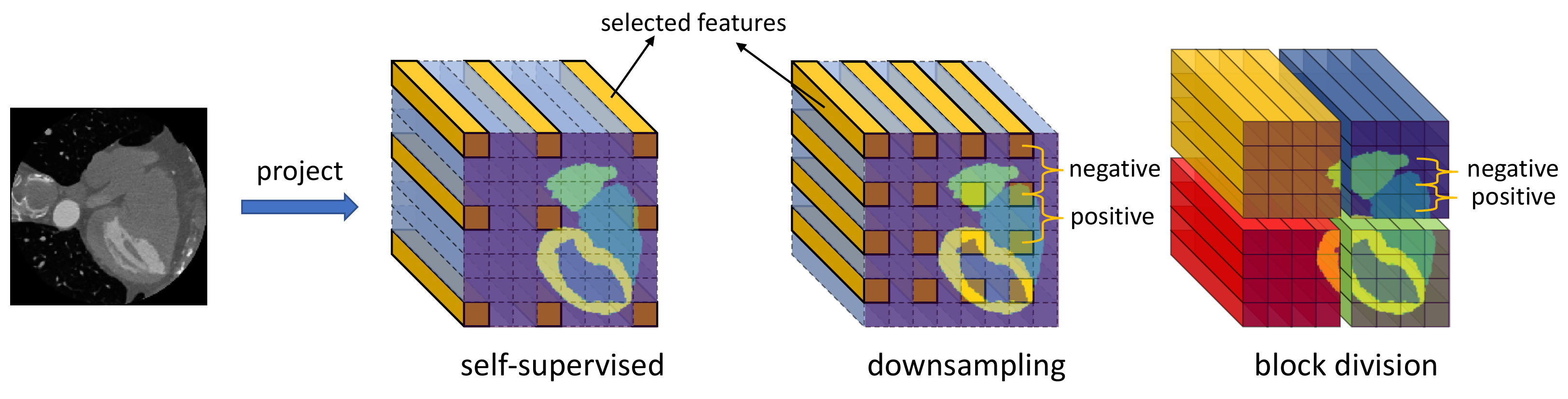}
    \caption{Illustration of three strategies to select features from feature maps for local contrastive learning. Left: self-supervised as in \cite{chaitanya2020contrastive}. Middle: downsampling where the contrastive loss is computed with features separated with fixed stride; Right: block division where contrastive loss is computed within each block and then averaged. }
    \label{fig:sample}
\end{figure}

As we try to utilize the segmentation annotation, the feature map is of the same size as the input image. 
Though we only consider the points that do not belong to background, $|P(u ,v)|$ can still be very large, that is $O(h^2)$ with input image dimension $h\times h$. The overall computation complexity for the superivsed local ontrastive loss is then $O(h^4)$. 
For example, the image size in MMWHS is $512\times512$. Even after we reshape the slices to $160\times160$, the size of negative set is on the order of $10^4$ and the total multiplications needed is on the order of $10^8$ to evaluate the loss. 
To address this issue, we propose two strategies to reduce the size of the negative set as illustrated in Fig.~\ref{fig:sample}.
The first one is to downsample the feature map with fixed stride.
The idea behind this method is that neighboring pixels contain redundant information and their embedded features appear quite similar in the latent space.
With stride number $s$, the size of $P(u,v)$ and the computation complexity of the loss function are reduced by a factor of $s^2$ and $s^4$ respectively.
As larger stride leads to fewer negative pairs and the number of negative pairs is important for contrastive learning\cite{he2020momentum}.
In this work, we set the stride to 4, the minimum value that will not cause out-of-memory (OOM) issue on our experiment platform. 

Alternatively, we can divide the feature maps into several blocks of the same size ${h}'\times{h}'$ as shown in Fig.~\ref{fig:sample}. The local contrastive loss is calculated within each block only and then averaged across all the blocks.
If all the pixels in a block are labeled as background, we will not compute its local contrastive loss.
The size of $P(u,v)$ decreases to the number of features in a block $O(h'^2)$ and the total computation complexity is $O((h/ h')^2 \cdot h'^4)=O(h^2h'^2)$.
Smaller block size will also lead to fewer negative pairs. Therefore the block size should also be determined based on the acceptable training time/memory consumption. 
In this paper, we set it as 16$\times$16, the maximum value that will not cause OOM.

\section{Experiments and Results}
\label{sec:experiment}
\textbf{Dataset and Preprocessing.} We evaluate our methods on two public medical image segmentation datasets of different modalities, MRI and CT. (1) Hippocampus dataset is from the MICCAI 2018 medical segmentation decathlon challenge\cite{Decathlon}. 
There are 260 3D volumes of mono-modal MRI scans with the annotation of anterior and posterior hippocampus body. 
(2) MMWHS \cite{zhuang2016multi} comes from the MICCAI 2017 challenge, which consists of 20 3D cardiac CT volumes.
The annotations include seven structures of heart: left ventricle, left atrium, right ventricle, right atrium, myocardium, ascending aorta, and pulmonary artery.
Regarding preprocessing, we normalize the intensity for the whole volume. Then we resize the slices of all volumes in the same dataset to unify them using bilinear interpolation.
The resolutions of the two reshaped dataset are $64\times64$ for Hippocampus and  $160\times 160$ for MMWHS.

\newcolumntype{Y}{>{\centering\arraybackslash}X}
\begin{table}
    \centering
    \caption{Comparison between state-of-the-art methods and the proposed methods w.r.t. subsequent segmentation dice scores on two datasets. Ablation studies of our methods (w/o global contrast) are also included.}
    \label{table1}
    \begin{tabularx}{0.9\textwidth}{c|Y|Y|Y|Y|Y|Y}
    \toprule
\multicolumn{1}{c|}{\multirow{3}{*}{Methods}}  & \multicolumn{3}{c|}{Hippocampus}  & \multicolumn{3}{c}{MMWHS} \\ \cline{2-7}
\multicolumn{1}{c|}{} & \multicolumn{3}{c|}{\% of data labeled in $X_{tr}$} & \multicolumn{3}{c}{\% of data labeled in $X_{tr}$} \\ \cline{2-7}
 \multicolumn{1}{c|}{} &\multicolumn{1}{c|}{5} & \multicolumn{1}{c|}{10} & \multicolumn{1}{c|}{20} & \multicolumn{1}{c|}{10}  & \multicolumn{1}{c|}{20} & \multicolumn{1}{c}{40}  \\
\hline
random      &\multicolumn{1}{c|}{0.788} & \multicolumn{1}{c|}{0.833}  & \multicolumn{1}{c|}{0.852} &\multicolumn{1}{c|}{0.328} &\multicolumn{1}{c|}{0.440} & \multicolumn{1}{c}{0.715}                \\
global\cite{chen2020simple} &\multicolumn{1}{c|}{0.817} & \multicolumn{1}{c|}{0.834}  & \multicolumn{1}{c|}{0.861} &\multicolumn{1}{c|}{0.359} &\multicolumn{1}{c|}{0.487} & \multicolumn{1}{c}{0.724}                  \\
global+local(self)\cite{chaitanya2020contrastive} &\multicolumn{1}{c|}{0.808} & \multicolumn{1}{c|}{0.843}  & \multicolumn{1}{c|}{0.858} &\multicolumn{1}{c|}{0.367} &\multicolumn{1}{c|}{0.490} & \multicolumn{1}{c}{0.730}       \\ 
Mixup\cite{zhang2017mixup}      &\multicolumn{1}{c|}{0.818} & \multicolumn{1}{c|}{0.847}  & \multicolumn{1}{c|}{0.861} &\multicolumn{1}{c|}{0.365} &\multicolumn{1}{c|}{0.541} & \multicolumn{1}{c}{0.755}                  \\
TCSM\cite{li2020transformation} &\multicolumn{1}{c|}{0.796} & \multicolumn{1}{c|}{0.838}  & \multicolumn{1}{c|}{0.855} &\multicolumn{1}{c|}{0.347} &\multicolumn{1}{c|}{0.489} & \multicolumn{1}{c}{0.733}                       \\ \hline
local(stride) &\multicolumn{1}{c|}{0.818} & \multicolumn{1}{c|}{0.845}  & \multicolumn{1}{c|}{0.860} &\multicolumn{1}{c|}{0.354} &\multicolumn{1}{c|}{0.485} & \multicolumn{1}{c}{0.743}          \\
local(block)  &\multicolumn{1}{c|}{0.817} & \multicolumn{1}{c|}{0.843}  & \multicolumn{1}{c|}{0.862} &\multicolumn{1}{c|}{0.366} &\multicolumn{1}{c|}{0.475} & \multicolumn{1}{c}{0.736}       \\
global+local(stride) &\multicolumn{1}{c|}{0.822} & \multicolumn{1}{c|}{\textbf{0.851}}  & \multicolumn{1}{c|}{0.863} &\multicolumn{1}{c|}{\textbf{0.384}} &\multicolumn{1}{c|}{0.525} & \multicolumn{1}{c}{0.758}       \\
global+local(block) &\textbf{0.824} &0.849 &\textbf{0.866} &0.382 &\textbf{0.553} & \textbf{0.764} \\
\bottomrule
\end{tabularx}
    
\end{table}

\textbf{Experiment Settings.} Similar to \cite{chaitanya2020contrastive}, the segmentation network backbone we choose is 2D U-Net with three blocks for both encoder and decoder path, the implementation of which is based on the PyTorch library.
For both sets, we divide the data into three parts, training set $X_{tr}$, validation set $X_{vl}$,  and test set $X_{ts}$. The ratio $|X_{tr}|:|X_{vl}|:|X_{ts}|$ for Hippocampus is 3:1:1 and 2:1:1 for MMWHS. 
Then we randomly choose certain number of samples in $X_{tr}$ as labeled volumes and the rest are viewed as unlabeled.  
We split the dataset four times independently to generate four different folds and the average dice score on $X_{ts}$ is used as the metric for comparison. 
We choose Adam as the optimizer for both contrastive learning and the following segmentation training. The learning rate is 0.0001 for contrastive learning and is 0.00001 for segmentation training.
The training epoch for contrastive learning is 70, and for segmentation training, it stops after 120 epochs. We run experiments of Hippocampus dataset on two NVIDIA GTX 1080 gpus and the experiments of MMWHS dataset on two NVIDIA Tesla P100 gpus.

We implement four variants of the proposed methods: the two-step global contrast and supervised local contrast with downsampling  (global+local(stride)) and block division (global+local(block); and for the purpose of ablation study, these two methods without global contrast (local(stride) and local(block)). 
For comparison, we implement two state-of-the-art contrast based methods: a global contrastive learning method~\cite{chen2020simple} (global), and a self-supervised global and local contrastive learning method~\cite{chaitanya2020contrastive} (global + local(self)).
Note that the entire training data is used but without any labels for global contrast and 
self-supervised local contrast, while only labeled training data is used for our supervised local contrast. 
We further implement three non-contrast baseline methods: training from scratch with random initialization (random), a data augmentation based method Mixup\cite{zhang2017mixup}, and a state-of-the-art semi-supervised learning method TCSM\cite{li2020transformation}.


\begin{figure}
    \centering
    \includegraphics[width=0.9\linewidth]{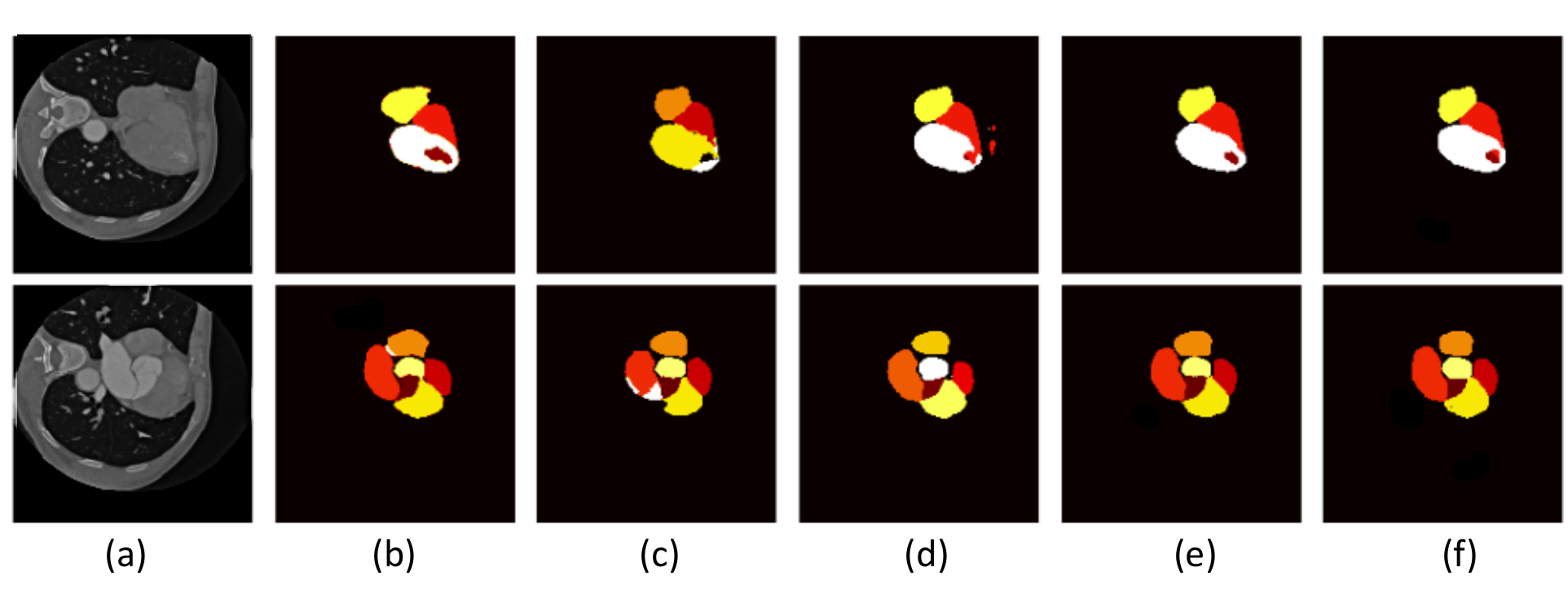}
    \caption{Visualization of segmentation results on MMWHS with 40\% labeled data. (a) and (b) represent input image and ground truth. (c) $\sim$ (f) are the predictions of random, global+local(self), global+local(stride), and global+local(block). Different color represents different structure categories. (Better to view in colors)}
    \label{fig:seg_result}
\end{figure}
Table \ref{table1} lists the average dice scores of all the methods. 
There are several important observations we can make from the table:
First, all contrastive learning methods lead to higher dice scores compared to the random approach, which implies that contrastive learning is a feasible initialization when labels are limited.
Second, we notice that the proposed supervised local contrast alone (local(stride) and local(block)) generates segmentation results comparable with those from existing contrast based methods. Considering that the former only makes use of the labeled data while the latter leverage all volumes, the supervised contrastive loss efficiently teaches the model the pixel-wise representations shared by the same class.
Lastly, combining the supervised local contrast with the global contrast, the semi-supervised setting gains a considerable improvement of segmentation performance over the state-of-the-art.
It demonstrates that the image-level representation extracted from unsupervised global contrast is complementary to the pixel-level representation from the supervised local contrast.

\begin{figure}
    \centering
    \includegraphics[width=0.9\linewidth]{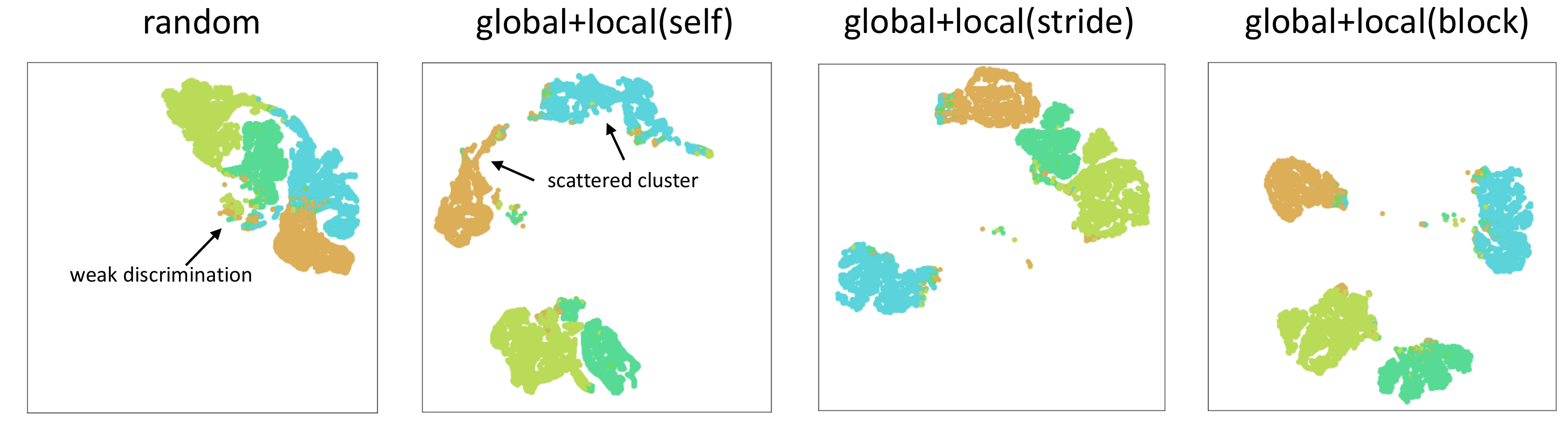}
    \caption{Visualization of embedding features after applying t-SNE on MMWHS with 40\% labeled data. Different color represents different classes. The features are all from the feature map before the last point-wise convolution. (Better to view in colors. }
    \label{fig:embedding}
\end{figure}

Finally, the segmentation results on MMWHS with 40\% labeled data from our methods, the state-of-the-art (global+local(self) \cite{chaitanya2020contrastive}), and the random approach are shown in Fig.~\ref{fig:seg_result}. From the figure we can clearly see that our approaches perform the best. With the same setting, the embedding features after applying t-SNE for these methods are visualized in Fig.~\ref{fig:embedding}. From the figure we can see that compared with the random approach, the features for the same class are tightly clustered and those for different classes are clearly separated for the global+local(block) method, while those from the state-of-the-art are more scattered. This aligns with the results in Table~\ref{table1} where global+local(block) method achieves the highest Dice (rightmost column). 

\section{Conclusion}
The limited annotation is always a hurdle for the development of deep learning methods for medical image segmentation.
Aiming at tackling this challenge, in this work, we propose a supervised local contrastive loss to learn the pixel-wise representation. 
Further, we put forward two strategies to reduce the computational complexity when computing the loss. Experiments on two public medical image datasets with only partial labels show that when combining the proposed supervised local contrast with global contrast, the resulting semi-supervised contrastive learning achieves substantially improved segmentation performance over the state-of-the-art. 

\section{Acknowledgement}
This work is partially supported by NSF award IIS-2039538.
\bibliographystyle{splncs04}
\bibliography{refs.bib}

\begin{thebibliography}{10}
\providecommand{\url}[1]{\texttt{#1}}
\providecommand{\urlprefix}{URL }
\providecommand{\doi}[1]{https://doi.org/#1}

\bibitem{Decathlon}
Medical segmentation decathlon chalenge.
  \url{http://medicaldecathlon.com/index.html} (2018)

\bibitem{bai2017semi}
Bai, W., Oktay, O., Sinclair, M., Suzuki, H., Rajchl, M., Tarroni, G., Glocker,
  B., King, A., Matthews, P.M., Rueckert, D.: Semi-supervised learning for
  network-based cardiac mr image segmentation. In: International Conference on
  Medical Image Computing and Computer-Assisted Intervention. pp. 253--260.
  Springer (2017)

\bibitem{bowles2018gan}
Bowles, C., Chen, L., Guerrero, R., Bentley, P., Gunn, R., Hammers, A., Dickie,
  D.A., Hern{\'a}ndez, M.V., Wardlaw, J., Rueckert, D.: Gan augmentation:
  Augmenting training data using generative adversarial networks. arXiv
  preprint arXiv:1810.10863  (2018)

\bibitem{chaitanya2020contrastive}
Chaitanya, K., Erdil, E., Karani, N., Konukoglu, E.: Contrastive learning of
  global and local features for medical image segmentation with limited
  annotations. arXiv preprint arXiv:2006.10511  (2020)

\bibitem{chaitanya2019semi}
Chaitanya, K., Karani, N., Baumgartner, C.F., Becker, A., Donati, O.,
  Konukoglu, E.: Semi-supervised and task-driven data augmentation. In:
  International conference on information processing in medical imaging. pp.
  29--41. Springer (2019)

\bibitem{chen2017deeplab}
Chen, L.C., Papandreou, G., Kokkinos, I., Murphy, K., Yuille, A.L.: Deeplab:
  Semantic image segmentation with deep convolutional nets, atrous convolution,
  and fully connected crfs. IEEE transactions on pattern analysis and machine
  intelligence  \textbf{40}(4),  834--848 (2017)

\bibitem{chen2020simple}
Chen, T., Kornblith, S., Norouzi, M., Hinton, G.: A simple framework for
  contrastive learning of visual representations. In: International conference
  on machine learning. pp. 1597--1607. PMLR (2020)

\bibitem{costa2017end}
Costa, P., Galdran, A., Meyer, M.I., Niemeijer, M., Abr{\`a}moff, M.,
  Mendon{\c{c}}a, A.M., Campilho, A.: End-to-end adversarial retinal image
  synthesis. IEEE transactions on medical imaging  \textbf{37}(3),  781--791
  (2017)

\bibitem{goodfellow2014generative}
Goodfellow, I.J., Pouget-Abadie, J., Mirza, M., Xu, B., Warde-Farley, D.,
  Ozair, S., Courville, A., Bengio, Y.: Generative adversarial networks. arXiv
  preprint arXiv:1406.2661  (2014)

\bibitem{he2020momentum}
He, K., Fan, H., Wu, Y., Xie, S., Girshick, R.: Momentum contrast for
  unsupervised visual representation learning. In: Proceedings of the IEEE/CVF
  Conference on Computer Vision and Pattern Recognition. pp. 9729--9738 (2020)

\bibitem{khosla2020supervised}
Khosla, P., Teterwak, P., Wang, C., Sarna, A., Tian, Y., Isola, P., Maschinot,
  A., Liu, C., Krishnan, D.: Supervised contrastive learning. arXiv preprint
  arXiv:2004.11362  (2020)

\bibitem{li2020shape}
Li, S., Zhang, C., He, X.: Shape-aware semi-supervised 3d semantic segmentation
  for medical images. In: International Conference on Medical Image Computing
  and Computer-Assisted Intervention. pp. 552--561. Springer (2020)

\bibitem{li2018h}
Li, X., Chen, H., Qi, X., Dou, Q., Fu, C.W., Heng, P.A.: H-denseunet: hybrid
  densely connected unet for liver and tumor segmentation from ct volumes. IEEE
  transactions on medical imaging  \textbf{37}(12),  2663--2674 (2018)

\bibitem{li2020transformation}
Li, X., Yu, L., Chen, H., Fu, C.W., Xing, L., Heng, P.A.:
  Transformation-consistent self-ensembling model for semisupervised medical
  image segmentation. IEEE Transactions on Neural Networks and Learning Systems
   (2020)

\bibitem{milletari2016v}
Milletari, F., Navab, N., Ahmadi, S.A.: V-net: Fully convolutional neural
  networks for volumetric medical image segmentation. In: 2016 fourth
  international conference on 3D vision (3DV). pp. 565--571. IEEE (2016)

\bibitem{ronneberger2015u}
Ronneberger, O., Fischer, P., Brox, T.: U-net: Convolutional networks for
  biomedical image segmentation. In: International Conference on Medical image
  computing and computer-assisted intervention. pp. 234--241. Springer (2015)

\bibitem{zeng2021positional}
Zeng, D., Wu, Y., Hu, X., Xu, X., Yuan, H., Huang, M., Zhuang, J., Hu, J., Shi,
  Y.: Positional contrastive learning for volumetricmedical image segmentation.
  arXiv preprint arXiv:2106.09157  (2021)

\bibitem{zhang2017mixup}
Zhang, H., Cisse, M., Dauphin, Y.N., Lopez-Paz, D.: mixup: Beyond empirical
  risk minimization. arXiv preprint arXiv:1710.09412  (2017)

\bibitem{zhang2017deep}
Zhang, Y., Yang, L., Chen, J., Fredericksen, M., Hughes, D.P., Chen, D.Z.: Deep
  adversarial networks for biomedical image segmentation utilizing unannotated
  images. In: International Conference on Medical Image Computing and
  Computer-Assisted Intervention. pp. 408--416. Springer (2017)

\bibitem{zhuang2016multi}
Zhuang, X., Shen, J.: Multi-scale patch and multi-modality atlases for whole
  heart segmentation of mri. Medical image analysis  \textbf{31},  77--87
  (2016)

\end{thebibliography}


\end{document}